\documentclass[11pt,a4paper]{article}
\usepackage[hyperref]{eacl2021}
\usepackage{times}
\usepackage{latexsym}

\usepackage{microtype}

\aclfinalcopy 

\newcommand{\rc}{{\sc ALBERT$_{QA}$}}
\newcommand{\mrc}{{\sc Mr.C}}
\newcommand{\eg}{\textit{e.g.}}

\newcommand{\SqSq}{\textbf{Train$_{SQ2}$-Test$_{SQ2}$}}
\newcommand{\SqNq}{\textbf{Train$_{SQ2}$-Test$_{NQ}$}}
\newcommand{\NqNq}{\textbf{Train$_{NQ}$-Test$_{NQ}$}}
\newcommand{\qpp}{QPP}
\usepackage{todonotes}
\definecolor{deepblue}{rgb}{0.16, 0.32, 0.75}
\definecolor{indiagreen}{rgb}{0.00, 0.44, 0.00}

\newcommand\tab[1][1cm]{\hspace*{#1}}
\usepackage{amsmath}
\usepackage{amssymb}
\usepackage{multirow}

\title{Towards Confident Machine Reading Comprehension}

\author{Rishav Chakravarti\thanks{\hspace{0.15cm}Work completed while at IBM Research AI} \\
  AWS AI Labs \\
  New York City, NY \\
  \texttt{chakrris@amazon.com} \\\And
  Avirup Sil\thanks{\hspace{0.15cm}Corresponding author} \\
  IBM Research AI \\
  Yorktown Heights, NY \\
  \texttt{avi@us.ibm.com} \\}

\date{}

\begin{document}
\maketitle
\begin{abstract}
There has been considerable progress on academic benchmarks for the Reading Comprehension (RC) task with State-of-the-Art models closing the gap with human performance on extractive question answering. Datasets such as SQuAD 2.0 \& NQ have also introduced an auxiliary task requiring models to predict when a question has no answer in the text. However, in production settings, it is also necessary to provide confidence estimates for the \textit{performance} of the underlying RC model at both answer extraction and ``answerability'' detection. We propose a novel post-prediction confidence estimation  model, which we call \mrc{} (short for Mr. Confident), that can be trained to improve a system's ability to refrain from making incorrect predictions with improvements of up to 4 points as measured by Area Under the Curve (AUC) scores. \mrc{} can benefit from a novel white-box feature that leverages the underlying RC model's gradients. Performance prediction is particularly important in cases of domain shift (as measured by training RC models on SQUAD 2.0 and evaluating on NQ), where \mrc{} not only improves AUC, but also traditional answerability prediction (as measured by a 5 point improvement in F1). 

\end{abstract}



\section{Introduction}
\label{sec:intro}

The reading comprehension (RC) task require models to extract or generate answers to questions about an input piece of text. Particularly since the advent of benchmark datasets such as SQuAD \cite{Rajpurkar_2016}, transfer learning models that leverage large pre-trained language models (LM) like BERT \cite{Devlin2018BERTPO} and ALBERT \cite{lan2019albert} have demonstrated high performance, even rivaling ``human'' performance.

\begin{figure}[!t]
    \centering
    \small
\framebox{%
  \begin{minipage}{0.95\columnwidth}
    \begin{center}
        \textbf{\underline{Contradictory Context}}
    \end{center}
        \textbf{Context:} Concrete hardens as a result of the chemical reaction between cement and water (known as hydration...For every pound (or kilogram or any unit of weight) of cement, \textbf{about 0.35 pounds} (or 0.35 kg or corresponding unit) of water is needed... However, \textbf{a mix with a ratio of 0.35 \textcolor{red}{may not mix thoroughly}}, and...ratios of \textbf{\textcolor{indiagreen}{0.45 to 0.60 are more typically used}}.\\
        \textbf{Question:} minimum required water cement ratio for a workable concrete is\\
        \textbf{Base RC Prediction:} \textcolor{red}{0.35}\\
        \textbf{Base RC Score:} 0.81\\
        \textbf{\mrc{} Score:} 0.19\\
        \textbf{Query Embedding Gradient Highlighting:} \\
        \tab{} \includegraphics[scale=0.45]{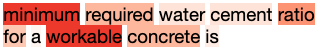}
    \begin{center}
        \textbf{\underline{RC Model Ignores Part of Question}}
    \end{center}
        \textbf{Context:} The \textbf{Inca Empire} at its greatest extent Capital Cusco (1438--1533)...Religion Inca religion Government Divine, \textbf{absolute monarchy}...\\ 
        \textbf{Question:} what government structure did the aztec and inca have in common\\
        \textbf{Base RC Prediction:} \textcolor{red}{absolute monarchy}\\
        \textbf{Base RC Score:} 0.94\\
        \textbf{\mrc{} Score:} 0.41\\
        \textbf{Query Embedding Gradient Highlighting:} \\
        \tab{} \includegraphics[scale=0.45]{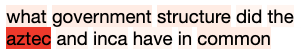}
  \end{minipage}}    
  \caption{Examples from the NQ dataset of bad answers from Base RC Model (in our case \rc{}) nonetheless produces high confidence scores. Our \mrc{} model produces lower confidence scores in these cases facilitating better thresholding.}
    \label{fig:examples}
\end{figure}

Despite this progress, the reliability of NLP models continues to be a concern with recent work demonstrating a lack of robustness to adversarial perturbations \cite{jia2017adversarial, wang2018robust,wallace-etal-2019-universal} as well as slight domain drift between training and inference \cite{niu2019automatically}. Despite these challenges, RC technologies are being adopted in industry settings (\eg{} Amazon Kendra\footnote{https://aws.amazon.com/kendra} offers users the ability to extract answers from their own corpora as part of its search offerings) where bad predictions can have adverse implications both in terms of reputation as well as actual harm \cite{hern_2017,klar_2019}.  

Recent benchmarks such as SQuAD 2.0 (SQ2) \cite{rajpurkar2018know} and Natural Questions (NQ) \cite{Kwiatkowski2019NaturalQA} have attempted to address reliability by introducing ``unanswerable'' questions into the datasets so as to force RC models to generate ``no answer'' predictions. SQ2 relies on crowd workers to generate both types of questions based on Wikipedia paragraphs while NQ organically collects questions from search logs and uses crowdworkers to annotate them as answerable or not based on Wikipedia articles.  However, this modification to the RC task only addresses the model's ability to recognize bad inputs (the traditional ``out-of-domain'' question detection \cite{Jia2020KnowWT,Tan2019OutofDomainDF}). While this is important, in production settings, it is also necessary to understand when the model performs poorly even in cases when an answer exists.  See Fig. \ref{fig:examples} highlighting examples from the NQ dataset where a (near) State-of-the-Art (SOTA) RC model fails to extract the correct answer, but makes a prediction nonetheless (with a high confidence score).  As we discuss in section \ref{sec:experiments}, such scenarios are particularly frequent when slight domain shifts take place (a common occurrence in real-life production scenarios).

This paper carries out a novel preliminary study on performance prediction where near SOTA RC models are evaluated on their ability to refrain from making incorrect predictions (both in the case of unanswerable questions as well as in the case where the model simply fails to make the right prediction despite its presence in the text). Using the SQ2 and NQ datasets, we find that:
\vspace{-0.8em}
\begin{enumerate}\setlength\itemsep{-0.5mm}
    \item A novel post-prediction confidence estimation  model, which we call \mrc{} (short for Mr. Confident), can be trained to improve a system's ability to refrain from making incorrect predictions with improvements of up to 4 points as measured by Area Under the Curve (AUC) scores.
    \item \mrc{} can benefit from a novel white-box feature that leverages the underlying RC model's gradients.
    \item  Performance prediction is particularly important in cases of domain shift (as measured by training RC models on SQ2 and evaluating on NQ), where \mrc{} not only improves AUC, but also traditional answerability prediction (as measured by a 5 point improvement in F1).
\end{enumerate}

\section{Related Work}
\label{sec:related}
\textbf{Answer Re-ranking:} 
\citet{wang2017evidence} aggregate evidences from different passages for open-doman QA and \citet{kratzwald2019rankqa} uses a combination of retrieval and comprehension features that are directly extracted from the
QA pipeline. \citet{nogueira2019passage} perform passage re-ranking using BERT which is another post-prediction task, but can suffer from the same challenge in production settings where the re-rank score is not necessarily informative for the purposes of thresholding ``good'' answers from ``bad'' ones. \\
\textbf{Answer Verification:} \citet{tan2018know,hu2019read+, zhang2019sg, zhang2020retrospective} perform answer validation and \citet{penas2007overview} adds an extra ``verifier" component to decide whether the predicted answer is entailed by the input snippets. \citet{back2020neurquri} perform requirement inspection by computing attention-based satisfaction score to compare question and candidate answer embeddings. Despite their similarity, these approaches continue to focus on discriminating between ``answerable'' and ``unanswerable'' questions without directly estimating the underlying RC model's prediction performance (regardless of answerability with respect to the context).  This work's focus on performance prediction (using metrics such as AUC) seems complimentary in nature to improvements in unanswerability detection. \\
\textbf{Query Performance Prediction (\qpp{}):} \citet{qpp2010, HE2006585,qpp2014} predict retrieval performance of document retrieval which aligns with our objective. However, as shown in our novel use of embedding gradients, there is room to introduce features and model types that are more tailored to the RC setting, which is not studied in the prior work.

\section{Method}
\label{sec:methods}

\subsection{Base RC Model Overview}
\label{subsec:rc}
To fulfill our objective to build a confident RC system, we start with a SOTA RC model as our base. Specifically, given a token sequence ($\mathbf{X}$) consisting of a question and a passage and special markers \eg{} the $[CLS]$ token for answerability classification,
the base RC model trains classifiers to predict the begin and end of answers spans as follows: $\boldsymbol{\alpha}_b = softmax(\mathbf{W}_1 \mathbf{H})$ and $\boldsymbol{\alpha}_e = softmax(\mathbf{W}_2 \mathbf{H})$  where $\mathbf{W}_1$, $\mathbf{W}_2 \in \mathbb{R}^{1\times D}$, where $\mathbf{H}$ is the contextualized representation of $\mathbf{X}$ provided by a deep Transformer \cite{Vaswani_2017} based language model (LM) and $D$ is the LM's output embedding dimension. An extra dense layer ($\mathbf{W}_3 \in \mathbb{R}^{5\times D}$) is added which operates only on the contextualized representation of the $[CLS]$ token to produce a likelihood prediction for additional answer types (required for NQ)\footnote{We only model long, short and null following prior work.}: $\boldsymbol{\alpha}_l^{[CLS]} = softmax(\mathbf{W}_3 \mathbf{h}^{[CLS]}))$. We choose ALBERT (\textit{xxlarge v2}) \cite{lan2019albert} as our LM and build our RC model, henceforth \rc{}, following \cite{zhang2020retrospective} for SQ2 and \cite{pan2019frustratingly,alberti2019bert} for NQ\footnote{At time of writing this was published single model SOTA.}. For more details, we direct interested readers to their papers and the appendix for hyperparameter settings.

\subsection{Confidence Estimation Model Overview}
\label{subsec:mrc}
\textbf{Learning Algorithm:} We introduce \mrc{}, a post prediction confidence estimation model, that utilizes a Gradient Boosted Machine (GBM) \cite{friedman2001} to learn an ensemble of weak learners (regression trees).  The objective function uses logistic regression loss to predict a binary label, $\mathbf{Y}$, indicating whether the max scoring answer span from the base \rc{} model correctly answers the question as per the official evaluation script's exact match criteria (a $NULL_SPAN$ indicator is expected for ``unanswerable'' questions). 

To derive training labels, we train \rc{} on a $90\%$ random split of the training data and generate predictions on the remaining $10\%$.  Both SQ2 and NQ are large datasets ($\sim130K$ query-passage pairs and $\sim300K$ query-article pairs respectively) so there is not a statistically significant deterioration going from a $100\%$ of the data to $90\%$ of the data. However, k-fold cross validation can be used to generate additional training data for \mrc{}.

\noindent\textbf{Input Features:} The following ``grey box'' features are used as input to \mrc{}:\vspace{-0.8em}
\begin{enumerate}\setlength\itemsep{-0.1em}
    \item \textbf{TopK-BeginLikelihood:} $\alpha_b$ for each of the top $k$ spans.
    \item \textbf{TopK-EndLikelihood:} $\alpha_e$ for each of the top $k$ spans.
    \item \textbf{Offset-Overlap-With-Top1:} the offset based F1 score between the predicted top 1 answer span and each of the remaining $k-1$ spans.
    \item \textbf{Query-IDF:} compute 4 features $[min, max, mean, skew]$ of the inverse document frequency scores for each of the input query tokens $q$ as computed on an English Wikipedia corpus.  
    \item \textbf{NoAnswer-Likelihood:} Compute $(\alpha_b^{[CLS]} + \alpha_e^{[CLS]})$. Typically, RC systems \cite{Devlin2018BERTPO} use this score to predict a ``no-answer'' as opposed to an answer string from the passage.
    \item \textbf{AnswerType-Likelihood:}  Compute $\alpha_l$ for all possible answer types (computed \textit{for NQ only}). 
\end{enumerate}
In addition, we derive a set of novel ``white box'' features: \textbf{Query-Embedding-Gradients} (QEG). We first use \rc{}'s top scoring $\alpha_b^{t_i} + \alpha_e^{t_j}$ offsets as the target labels to compute cross entropy loss. This loss is back propagated to compute the gradients for each of the input query token embeddings (i.e. the input of the underlying ALBERT LM)\footnote{We take the $L^2$ norm to compute a single value per token}.   Finally, we compute $[min, max, mean, skew]$ over each of the token embedding gradient norms.

\noindent\textbf{Intuition:} Features 1--4 are inspired by previous work in QPP \cite{qpp2010}: 1--3 derive signals   from the top answer and potential other answers as a post retrieval scoring mechanism, whereas 4 derives inspiration from specificity which tend to favor specific queries over general ones.  Features 5 \& 6 track ``unanswerability'' prediction. Finally, QEG looks at how well aligned the underlying query terms were with the context as seen by \rc{}.



\section{Experiments}
\label{sec:experiments}



\noindent\textbf{Dataset + Conditions:} We evaluate \mrc{} on these datasets under three different training conditions:\\
    \noindent \textbf{1.} \SqSq{}\textbf{:} Train on SQ2 + evaluate on SQ2.\\
    \noindent \textbf{2.} \SqNq{}\textbf{:} Train on SQ2 + evaluate on NQ\: Short Answer\footnote{SQ2 does not have long answer selection, so we provide the model with the oracle long answer paragraphs.}. We use this as a measure model robustness since both tasks are fairly similar. \\
    \noindent \textbf{3.} \NqNq{}\textbf{:} Train on NQ + evaluate on NQ: Short and Long Answers\\

\noindent\textbf{Evaluation Metric:} Rather than focusing purely on the official F1 measure evaluated at an optimally chosen threshold to balance precision on ``answerable'' questions while limiting recall of ``unanswerable'' questions, we evaluate using area under the receiver operating characteristics (ROC) curve (AUC) measuring model prediction correctness (regardless of ``answerability'') to summarize performance across a variety of threshold settings \cite{auc}.  In all cases, we evaluate on the official dev sets since this requires question level model performance metrics which are inaccessible for the test sets\footnote{Leaderboard results show that dev generalizes to test}.

\begin{table}
\small
\begin{center}
\resizebox{0.95\columnwidth}{!}{\begin{tabular}{|l||c|c|c|c|}
\hline
\multirow{2}{*}{Train on SQ2} &
\multicolumn{2}{c|}{SQ2} & 
\multicolumn{2}{c|}{NQ} \\
\cline{2-5}
 & F1 & AUC & F1 & AUC\\
\hline
\hline
\cite{zhang2020retrospective} & $88.8$ & - & - & -  \\
\hline
\hline
\rc{} & $89.1$ & $81.1$   & $62.0$ & $55.4$  \\
\hline
 + \mrc{} & $88.7$ & $\mathbf{85.0}$ & $\mathbf{68.2}$ & $\mathbf{65.9}$\\
\hline \multicolumn{5}{}{}\\
\hline
\multirow{2}{*}{Train on NQ} & 
\multicolumn{2}{c|}{Long Answer} & 
\multicolumn{2}{c|}{Short Answer} \\
\cline{2-5}
 & F1 & AUC & F1 & AUC \\
\hline
\hline
\cite{pan2019frustratingly} & $68.2$ & - & 57.2 & - \\
\hline
\hline
\rc{} & $71.0$ & $85.4$ & $60.1$ & $92.5$ \\
\hline
 + \mrc{} & $\mathbf{72.8}$ & $\mathbf{88.0}$ &  $59.8$ & $92.6$  \\
\hline
\end{tabular}}
\end{center}
\caption{Bold denotes statistically significant differences. We also compare against the prior published SOTA. F1 scores are calculated using the official evaluation scripts for SQ2 and NQ.} 
\label{tab:results}
\end{table}
\noindent\textbf{Results:} Table \ref{tab:results} show gains in AUC over \rc{} (denoted as Base in the tables) across all train and dev configurations with the largest improvement being in the \SqNq{} setting suggesting that \mrc{} generalizes better (perhaps due to more robust features or simpler modeling assumptions). The AUC increases are statistically significant ($\alpha=0.01$ based on a bootstrap randomization test) with the exception of the Short Answer setting in \NqNq{}. The NQ Short Answer subset is skewed towards ``unanswerable'' questions and it seems to do a reasonably good job at learning to identify these questions. However, it is still failing to detect prediction errors on ``answerable'' questions (we provide a breakdown in the appendix where we see that the AUC on the answerable subset is 2.7 points higher for \mrc{}). This suggests future work to extract features which better leverage the NQ trained \rc{} model's ability to detect unanswerable questions.

In addition, the official NQ script's exact match is stricter than that of SQ2. SQ2 allows slight deviations (e.g. casing) and correct tokens can be predicted from any context within the passage.  During evaluation, multiple annotations per example are available in the NQ dev set (so the answer can match any of the variations), but the train set (used to provide labels for \mrc{} training) only contains a single annotation per example. As future work, we intend to explore softer matches to encourage \mrc{}'s hill climbing during the training process.


There is a slight drop in the base F1 measurement (which only looks at ``un-answerability'' and not on performance prediction) on \SqSq{} as well as the Short Answer portion of \NqNq{}.  However, neither of these drops are statistically significant and, in fact, there is a dramatic and statistically significant F1 increase in the \SqNq{} setting.  So overall, \mrc{} tends to be a more robust decider for whether to trust the underlying \rc{} model prediction.

\noindent\textbf{Feature analysis:} We also analyze the informativeness of the input features for \mrc{} and find that the $mean$ \textbf{QEG} feature is the most influential feature in the \SqNq{} and \SqSq{} (as measured by the average decrease in mean squared error over tree nodes involving this feature in the GBM). Refer to the appendix for a full list of features sorted by their relative influence.

During manual error analysis on a random sample, we also find that this feature appears to provide a useful visualization of the errors with respect to the query terms \eg{} Fig \ref{fig:examples} shows the query tokens highlighted based on their \textbf{QEG} values.  The terms with the largest gradients do seem like the critical ones which are least aligned with the evidence in the context so may provide useful insight to the reader.
\section{Conclusion}
Making errors in academic benchmarks results in a lower F1 score but making errors \textit{confidently} in production systems is, at best, embarrassing \cite{klar_2019} and, at worst, harmful \cite{hern_2017}. We propose \mrc{}, a post prediction confidence estimation model that maintains the base system accuracy, while providing statistically significantly better judgments for the RC model prediction's accuracy. Previous work in RC has only addressed either re-ranking or detecting ``un-answerable'' questions rather than directly modeling the prediction error of the base RC system: we hope this novel work in RC will spur a future research direction and lead to more confident RC systems.

\bibliographystyle{acl_natbib}
\bibliography{references}

\section*{Appendix}

\subsection*{\rc{} Model Training}

The \rc{} training was carried out using the pytorch transformers toolkit made available by Huggingface\footnote{\url{https://github.com/huggingface/transformers}}.  Models were trained and inference using single GPU training on machines with 32GB Tesla V100 GPUs with 16-bit precision. This results in processing speeds of $\sim{70}$ features per second. Note that, in line with prior work on using transformer models in the MRC setting \cite{Devlin2018BERTPO}, a single query-context pair can be split into multiple features in order to accommodate the maximum sequence length.

\subsubsection*{SQuAD 2.0 (SQ2) Training}
Table \ref{tab:sq2hyp} lists the hyperparameters used in line with \citet{lan2019albert} for training \rc{} on SQ2.

\begin{table}[h]
\begin{tabular}{| l | c |}
\hline
\textbf{Hyperparameter} & \textbf{Setting}\\ \hline \hline
Max query tokens & 64 \\ \hline
Max answer tokens & 30 \\ \hline
Batch size & 32  \\ \hline
Optimizer & Adam \\
 $\epsilon$ & $1^{-8}$\\ 
 $\beta_1$ & $0.9$\\ 
 $\beta_2$ & $0.999$\\ \hline
Warmup ratio \% & 10\%   \\ \hline
Learning rate warmup & Linear \\ \hline
Peak learning rate & 3e-5  \\ \hline
Epochs & 3 \\ \hline
Document stride & 128 \\ \hline
Max sequence length & 512 \\ \hline
\end{tabular}
\caption{SQ2 \rc{} Hyperparameter Configurations}
\label{tab:sq2hyp}
\end{table}

\subsubsection*{Natural Questions (NQ) Training}
Table \ref{tab:nqhyp} lists the hyperparameters used in line with \citet{pan2019frustratingly,alberti2019bert} for training \rc{} on NQ.

\begin{table}[h]
\begin{tabular}{| l | c |}
\hline
\textbf{Hyperparameter} & \textbf{Setting}\\ \hline \hline
Max query tokens & 18 \\ \hline
Max HTML spans & 48 \\
(top level) & \\ \hline
Max answer tokens & 30 \\ \hline
Batch size & 48  \\ \hline
Optimizer & Adam \\
 $\epsilon$ & $1^{-8}$\\ 
 $\beta_1$ & $0.9$\\ 
 $\beta_2$ & $0.999$\\ \hline
Warmup ratio & $10\%$   \\ \hline
Learning rate warmup & Linear \\ \hline
Peak learning rate & 1.6E-5  \\ \hline
Weight decay & 0.01  \\ \hline
Gradient clipping (norm) & 1.0\\ \hline
Epochs & 1 \\ \hline
Document stride & 192 \\ \hline
Max sequence length & 512 \\ \hline
Negative Subsampling  & $4\%$ \\
(Answerable Questions) & \\ \hline
Negative Subsampling  & $1\%$ \\
(Un-answerable Questions) & \\ \hline
\end{tabular}
\caption{NQ \rc{} Hyperparameter Configurations}
\label{tab:nqhyp}
\end{table}

\subsection*{\mrc{} Model Training}

The \mrc{} model training uses R's gbm implementation\footnote{\url{https://cran.r-project.org/package=gbm}} of the Gradient Boosted Machine (GBM). Training and inference was carried out on a 2.7 GHz Intel Quad Core i7 processor.

\textbf{Note about target labels:} The target function uses logistic regression loss where the target label, $\mathbf{Y}$, is a binary label indicating whether or not the max scoring answer span from the base \rc{} model correctly answered the question as measured by exact match. We use the official SQuAD 2.0 evaluation script's definition\footnote{\url{https://rajpurkar.github.io/SQuAD-explorer/}} of ``exact match'' for the SQuAD results (which is actually a fuzzy token match) and the NQ official evaluation script's definition\footnote{\url{https://github.com/google-research-datasets/natural-questions/}} of exact match for the NQ results (which is an exact span offset match).

Before training the final \mrc{} model on these features, we perform (1) feature selection (based on an information gain thresholding criteria) and (2) hyper-parameter selection via grid-search using a 75-25 split of the training data.

\subsubsection*{Feature Selection}
To perform feature selection we:
\begin{enumerate}
    \item impute a random noise feature (by sampling from a uniform distribution between $[0-1]$)
    \item train a model on the $75\%$ split of the data
    \item evaluate on the $25\%$ split while calculating an information gain based metric for each feature by averaging the error reductions from all splits in the regression tree that use that feature.  This is provided as ``relative.influence'' by the gbm toolkit.
    \item drop any features that had an influence score less than that of the random feature.
\end{enumerate}

Overall 22 features were selected for the model trained using the SQ2 \rc{} model's predictions (top 10 most informative features are summarized in table \ref{tab:sqfeatures}) and 18 were selected for the model trained using the NQ \rc{} model's predictions  (top 10 most informative features are summarized in table \ref{tab:nqfeatures}).

\begin{table}[]
\begin{tabular}{| l | l |}
\hline
Feature                           & \multicolumn{1}{c |}{Average Reduction} \\
&  \multicolumn{1}{c |}{in Error} \\
\hline
\hline
$mean(\mathbf{QEG})$  & 23.32                      \\
\hline
$\alpha_e$\_$top_1$     & 16.30                      \\
\hline
$\alpha_b$\_$top_1$   & 11.38                      \\
\hline
$\alpha_e$\_$top_2$     & 6.00                       \\
\hline
$min(\mathbf{QEG})$   & 5.86                       \\
\hline
$\alpha_e$\_$top_3$     & 4.12                       \\
\hline
$f1$--$overlap(top_1, top_3)$ & 3.85                       \\
\hline
$\alpha_b$\_$top_3$   & 3.55                       \\
\hline
$\alpha_b$\_$top_2$   & 3.27                       \\
\hline
$\alpha_b$\_$top_5$   & 3.11                  \\
\hline
\end{tabular}
\caption{Top 10 most influential \mrc{} features for SQ2. $\mathbf{QEG}$ = Query Embedding Gradient feature.}
\label{tab:sqfeatures}
\end{table}

\begin{table}[]
\begin{tabular}{| l | l |}
\hline
Feature                           & \multicolumn{1}{c |}{Average Reduction} \\
&  \multicolumn{1}{c |}{in Error} \\ \hline \hline
$\alpha_b + \alpha_e$\_$top_1$       & 74.87                      \\ \hline
$\alpha_b^{[CLS]} + \alpha_e^{[CLS]}$\_$top_1$                   & 8.47                       \\ \hline
$\alpha_b + \alpha_e$\_$top_2$       & 2.32                       \\ \hline
$\alpha_b + \alpha_e$\_$top_4$       & 1.94                       \\ \hline
$\alpha_{l==short}top_4$ & 1.60                       \\ \hline
$\alpha_b^{[CLS]} + \alpha_e^{[CLS]}$\_$top_2$                   & 1.25                       \\ \hline
$\alpha_{l==short}top_3$ & 1.12                       \\ \hline
$\alpha_b + \alpha_e$\_$top_5$       & 1.11                       \\ \hline
$\alpha_b + \alpha_e$\_$top_3$       & 0.97                       \\ \hline
$f1$--$overlap(top_1, top_4)$        & 0.94                     \\ \hline 
\end{tabular}
\caption{Top 10 most influential \mrc{} features for NQ}
\label{tab:nqfeatures}
\end{table}

\begin{table}[]
\begin{tabular}{| l | l |}
\hline
\textbf{Hyperparameter} & \textbf{Setting}\\ \hline \hline
Learning rate                & 0.1 \\ \hline
Interaction depth            & 3   \\ \hline
Minimum observations in node & 100 \\ \hline
Bag fraction                 & 1   \\ \hline
Number of trees              & 207  \\ \hline
\end{tabular}
\caption{Final hyperparameters for \mrc{} NQ}
\label{tab:mrc-nq-hyp}
\end{table}

\begin{table}[]
\begin{tabular}{|l|l|}
\hline
\textbf{Hyperparameter} & \textbf{Setting}\\ \hline \hline
Learning rate                & 0.05 \\ \hline
Interaction depth            & 4   \\ \hline
Minimum observations in node & 100 \\ \hline
Bag fraction                 & 0.8   \\ \hline
Number of trees              & 468 \\ \hline
\end{tabular}
\caption{Final hyperparameters for \mrc{} SQ2}
\label{tab:mrc-sq2-hyp}
\end{table}

After feature selection a grid search was performed over the hyperparameters again using the $75-25$ split yielding the hyperparameters in table \ref{tab:mrc-sq2-hyp} for SQ2 and those in table \ref{tab:mrc-nq-hyp} for NQ. These parameters are then applied for a final training on the full dataset (note that the ``full'' data set here is still only $10\%$ of the original SQ2 or NQ training data).

\subsection*{Additional Results}

In table \ref{tab:hasans}, we break down the results of the AUC just on the ``answerable'' subset of questions for each data set.

\begin{table}[]
\begin{tabular}{|l|l|l|}
\hline
\multicolumn{3}{|c|}{SQ2}                          \\ \hline
\hline
                          & AUC  & AUC$_{answerable}$\\ \hline
\rc{}   & 81.1 & 97.3            \\ \hline
+ \mrc{} & 85.0   & 95.9            \\ \hline
\hline
\multicolumn{3}{|c|}{NQ: Long Answer}              \\ \hline \hline
                          & AUC  & AUC$_{answerable}$\\ \hline
\hline
\rc{}    & 85.4 & 70              \\ \hline
+ \mrc{} & 88.0   & 77.9            \\ \hline
\hline
\multicolumn{3}{|c|}{NQ: Short Answer}             \\ \hline
\hline
                          & AUC  & AUC$_{answerable}$\\ \hline
                          
\rc{}   & 92.5 & 82.5            \\ \hline
+ \mrc{} & 92.6 & 85.2            \\ \hline
\end{tabular}
\caption{AUC broken down into the full dataset and just the ``answerable'' subset}
\label{tab:hasans}
\end{table}
\end{document}